\documentclass[letterpaper, 10 pt, conference]{ieeeconf}  

\IEEEoverridecommandlockouts                              

\usepackage{graphics} 
\usepackage{epsfig} 
\usepackage{mathptmx} 
\usepackage{times} 
\usepackage{amsmath} 
\usepackage{amssymb}  
\usepackage[ruled,vlined]{algorithm2e}
\usepackage{multirow}
\usepackage{color}

\usepackage{url}
\usepackage{hyperref}
\usepackage{booktabs}

\title{\LARGE \bf
Safe Reinforcement Learning using Formal Verification for Tissue Retraction in Autonomous Robotic-Assisted Surgery
}

\author{Ameya Pore$^{1,2,+}$,  Davide Corsi$^{1,+}$, Enrico Marchesini $^{1,+}$, Diego Dall'Alba$^{1}$, \\ Alicia Casals$^{2}$, Alessandro Farinelli$^{1}$ and Paolo Fiorini$^{1}$
\thanks{$^{+}$ Equal contribution} 
\thanks{$^{1}$ Department of Computer Science, University of Verona, Verona, Italy
        {\tt\small }}%
\thanks{$^{2}$ Center of Research in Biomedical Engineering, Universitat Politècnica de Catalunya , Barcelona, Spain
        {\tt\small}}%
\thanks{corresponding author: 
        {\tt\small ameya.pore@univr.it}}%
\thanks{This project has received funding from the European Union’s Horizon 2020 research and innovation programme under the Marie Skłodowska-Curie (grant agreement No. 813782 "ATLAS") and under (grant agreement No. 742671 "ARS")}
}

\begin{document}

\maketitle
\thispagestyle{empty}
\pagestyle{empty}

\begin{abstract}
Deep Reinforcement Learning (DRL) is a viable solution for automating repetitive surgical subtasks due to its ability to learn complex behaviours in a dynamic environment.
This task automation could lead to reduced surgeon’s cognitive workload, increased precision in critical aspects of the surgery, and fewer patient-related complications.
However, current DRL methods do not guarantee any safety criteria as they maximise cumulative rewards without considering the risks associated with the actions performed. Due to this limitation, the application of DRL in the safety-critical paradigm of robot-assisted Minimally Invasive Surgery (MIS) has been constrained. In this work, we introduce a Safe-DRL framework that incorporates safety constraints for the automation of surgical subtasks via DRL training. We validate our approach in a virtual scene that replicates a tissue retraction task commonly occurring in multiple phases of an MIS. Furthermore, to evaluate the safe behaviour of the robotic arms, we formulate a formal verification tool for DRL methods that provides the probability of unsafe configurations. Our results indicate that a formal analysis guarantees safety with high confidence such that the robotic instruments operate within the safe workspace and avoid hazardous interaction with other anatomical structures.

\end{abstract}

\section{INTRODUCTION}
Tissue retraction (TR) is a recurring subtask carried out during a Minimally Invasive Surgery (MIS) that involves manipulating deformable connective tissues to access the region of interest such as a tumour and takes a significant time during each procedure \cite{steele2013current}. Nowadays, MIS is commonly assisted via robotic platforms such as the DaVinci Surgical System (DVSS) that consist of several instruments. DVSS comprises three robotic arms called Patient Side Manipulator (PSM) equipped with articulated MIS instruments and controlled by the surgeon via a console endowed with two master handlers. Robot-assisted MIS may require that TR is either temporarily carried out using the third PSM or additional instruments are used that are controlled by an assistant operator  \cite{attanasio2020autonomous}. Since TR is one of the frequently used surgical gestures in multiple phases of surgery, it requires the surgeon to continuously switch between robotic arms or coordinate with the assistant operator. This protocol significantly increases the cognitive load of the surgeon and the risks of tissue damage. Hence, automation of the TR subtask can benefit surgeons by allowing them to focus on critical aspects of the surgery and potentially improve the overall outcome.

\begin{figure}[t]
	\centering
	\includegraphics[width=0.49\textwidth]{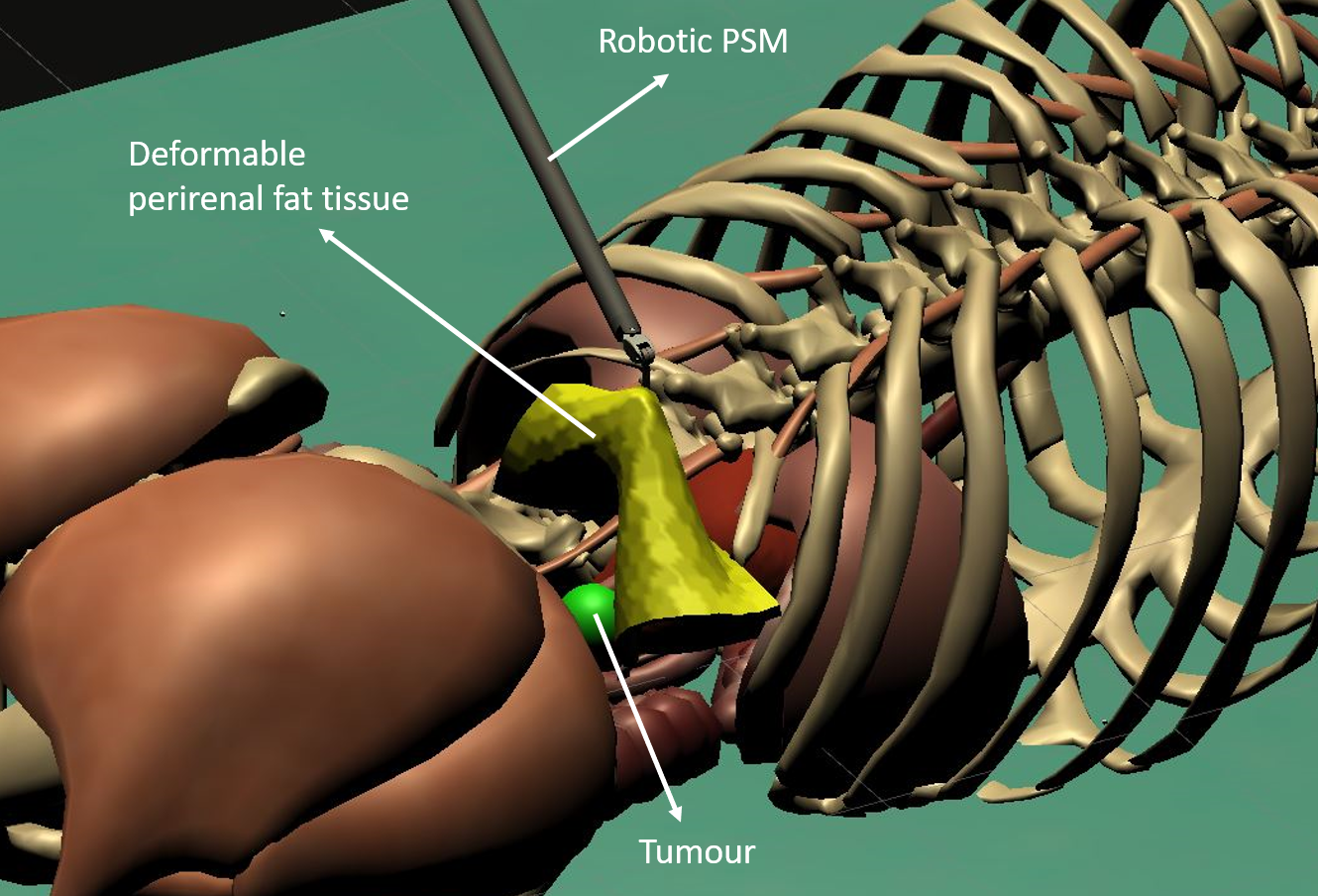}
	\caption{The virtual scene used to simulate the tissue retraction task during a partial nephrectomy procedure. The yellow tissue represents the renal adipose tissue that needs to be retracted to expose the tumour (green sphere) embedded in the underlying kidney (not visible in the picture).}
	\label{fig:scene}
\end{figure}

Recent efforts in surgical subtask automation have shown a surge in interest in employing data-driven approaches such as Deep Reinforcement Learning (DRL) \cite{richter2019open, nguyen2019manipulating}. 
DRL has provided breakthrough success in other robotic domains, namely manipulation \cite{levine2018learning}, navigation and locomotion tasks \cite{marchesini2020navigation}. 
However, the training of DRL methods is based on the exploration of state space efficiently and do not implicitly consider the risk associated with the actions \cite{garcia2015comprehensive}. DRL methods find an optimal policy by maximising long-term rewards but do not avoid the rare occurrence of a large negative reward that often corresponds to high-risk actions. 
Training DRL methods using a virtual environment is a widely adopted practice due to their data-hungry training regime that requires a high number of trial and error attempts.
This is even more prominent in the case of robot-assisted MIS, where strict ethical, legal and economic constraints require training and validating automation methods in a simulated environment before applying them in a real scenario.
Recent works have proposed surgical simulation environments suitable to train DRL methods \cite{richter2019open, tagliabue2020soft}. However, utmost safety concerns with DRL methods have limited their deployment in a clinical setting.
The guarantee of a provable behaviour using DRL is still an open problem and is a prime concern in their universal application for building trustworthy solutions \cite{verification_survey}. 

A recent research direction that addresses safety in DRL aims at incorporating auxiliary objectives to address the risk associated with actions. Multi-Objective Reinforcement Learning (MORL) and Constrained Reinforcement Learning (CRL) learn policies to simultaneously optimise several criteria \cite{yang2019morl} or limit the accumulation of such auxiliary functions \cite{Ray2019safetygym}. These approaches model risks via a cost function, e.g. measuring the cost as the number of collisions. However, they typically result in poor performance and policies that are not of high-quality \cite{Ray2019safetygym}. A more intuitive solution to improve safety is via reward shaping \cite{garcia2015comprehensive}, which can be naturally incorporated in well-defined training procedures such as the one we consider in our work.
DRL methods use Deep Neural Network (DNN) that can show unanticipated behaviour if the input data comes outside the training regime; hence ensuring that the network never makes decisions that can cause a property violation is crucial. Such validation requires estimating violations without executing the network, i.e. without performing the actions in a DRL setup. Running the network over many experiments and counting the unsafe configurations can be time-consuming and can only give an empirical evaluation without any guarantee of safety \cite{verification_survey}. Moreover, existing methods do not offer metrics to examine the level of safety offered by autonomous control algorithms. For this reason, we formulate a formal verification tool that allows us to mathematically guarantee the safety of the learnt behaviours with respect to pre-defined safety rules, referred to as properties. Furthermore, we define a metrics, called violation rate, that enables an evaluation of how often a trained DRL model (with rarely occurring small adversarial perturbations) will violate the properties. 

In summary, we introduce a Safe-DRL framework for the automation of the TR surgical subtask. We formulate the safety problem for TR as a set of properties that provide the limits of the safe workspace, such that the PSM does not collide with the surrounding anatomical structures. To evaluate the safety, we provide a formal verification analysis that gives us a probability of unsafe configurations over the designed set of properties. Our experimental scene consists of a virtual environment for a robot-assisted partial nephrectomy procedure that extensively requires manipulation and TR of perirenal fat tissue that covers the kidney to expose the region of interest (see Fig.~\ref{fig:scene}). One of the challenges in automating the TR task is to account for the heterogeneous and dynamic properties of the deformable fat tissue without disrupting the nearby tissues \cite{li2020super}. Hence, our contribution is twofold: first, introducing a safety framework for automating surgical subtasks for DRL methods and second, providing a formal verification tool for evaluating the violation of safety properties.

The outline of this paper is as follows: In Sec.~\ref{relatedworks}, we study related works in the paradigm of Safe-DRL methods and the recent application of DRL in surgical procedures. We explain our methods in Sec.~\ref{methods}, while in Sec.~\ref{experiments}, we elaborate on the experiments carried out for their validation. We present the results obtained in Sec.~\ref{resutls}. Finally, we discuss the implication of this work and subsequent direction in Sec.~\ref{conclusion}. 

\section{RELATED WORKS} \label{relatedworks}

TR is a relatively less studied subtask, although it frequently occurs in surgical scenarios. One of the earliest works in TR utilised probabilistic road-maps for optimisation based objectives \cite{patil2010toward}. Nagy et al. developed a TR algorithm based on images using hidden Markov models with fuzzy logic \cite{nagy2018surgical}. Recently, Attanasio et al. proposed a TR trajectory planner that is based on the coordinates extracted from images using a visual model \cite{attanasio2020autonomous}. The above methods use hand-engineered motion sequences; hence demonstrating complex non-linear behaviours may be challenging.
 
\textit{DRL for soft-tissue manipulation}: Previous works used a DRL based approach to learn a tensioning policy for surgical soft tissue multiple pinch point cutting task \cite{thananjeyan2017multilateral,nguyen2019manipulating}. Shin et al. used a Reinforcement Learning (RL) based approach to learn model predictive control for tissue dynamics \cite{shin2019autonomous}, while Pedram et al. used handcrafted features to incorporate prior knowledge in a vision-based RL approach \cite{pedram2019toward}. Richter et al. introduced a virtual environment in which model-free DRL can be trained with static surgical objects \cite{richter2019open}. Alternatively, to increase the sample-efficiency of DRL methods and obtain near-human behaviours, some studies have used imitation learning methods that require learning from demonstrations \cite{murali2015learning, chi2020collaborative}. Recently, Tagliabue et al. proposed a framework, \textit{UnityFlexML} in which deformable soft-tissue can be simulated and trained using DRL methods \cite{tagliabue2020soft}.
Note that none of these studies considers safety constraints to demonstrate the required behaviour.

\textit{Safe RL}: Several learning techniques have been recently proposed to incorporate auxiliary objectives in the training process and improve safety. MORL aims at optimising an additional cost function that is designed to measure safety \cite{yang2019morl}. Explicitly learning behaviours over multiple objectives is challenging and prior work either converge to an average policy \cite{Vamplew2011morl}, or present scalability issues \cite{yang2019morl}. Similarly, CRL \cite{Ray2019safetygym} has been used to introduce safety constraints in the training phase by limiting the accumulation of the cost function. These approaches, however, result in a significant trade-off in functional performance as the constraints severely limit the exploration process, negatively influencing the learned behaviours. A more intuitive way to address safety in well-defined tasks, such as the one we consider in our work, is via reward shaping \cite{garcia2015comprehensive}. As detailed in Section \ref{methods}, the idea is to exploit domain knowledge to design proxy reward functions that lead the trained policy to perform the desired safe behaviours.

\textit{Formal Verification of Neural Networks}:
Evaluation of the robust nature of DNNs is an open problem and can be addressed in many ways \cite{robotics_verification}. One of the first approaches, ReluPlex \cite{reluplex} proposed to find the largest neighbourhood around a point in feature space that guarantees that no point inside this area can change the classifier decision (i.e., small perturbations in the input does not change the network decision). However, such verification is NP-complete and does not scale well with huge input spaces \cite{np_complete}. A new family of formal analysis for DNNs has been adopted using interval algebra \cite{interval_algebra} to verify handcrafted safety properties \cite{prove}. FastLin \cite{fastlin} exploited the linear approximation of the ReLU units to provide an efficient and scalable algorithm, while Neurify \cite{neurify} relied on the symbolic interval analysis to give a strict estimation of the output bounds from a subset of the input space. However, all these methods can not be easily adapted to a reinforcement learning scenario, where a network encodes a sequential decision making problem and can not provide metrics to evaluate the safety of such models. In the next section, we show how we adapt the standard approaches to our context, presenting novel metrics specifically designed to evaluate a policy from a safety perspective.


\section{METHODS}\label{methods}
Our objective in this work is to accomplish the TR task, i.e. expose the underlying tumour without interacting with the nearby tissues or organs. For that, 
we consider a laparoscopic surgical scene that consists of the DVSS robotic PSM and several organs with a kidney covered by a perirenal fat tissue (see Fig.~\ref{fig:scene}). We use the 
\textit{UnityFlexML} framework to simulate the deformable fat tissue \cite{tagliabue2020soft}. 

\begin{figure}[t]
	\centering
    \vspace{.5em}
	\includegraphics[width=0.49\textwidth]{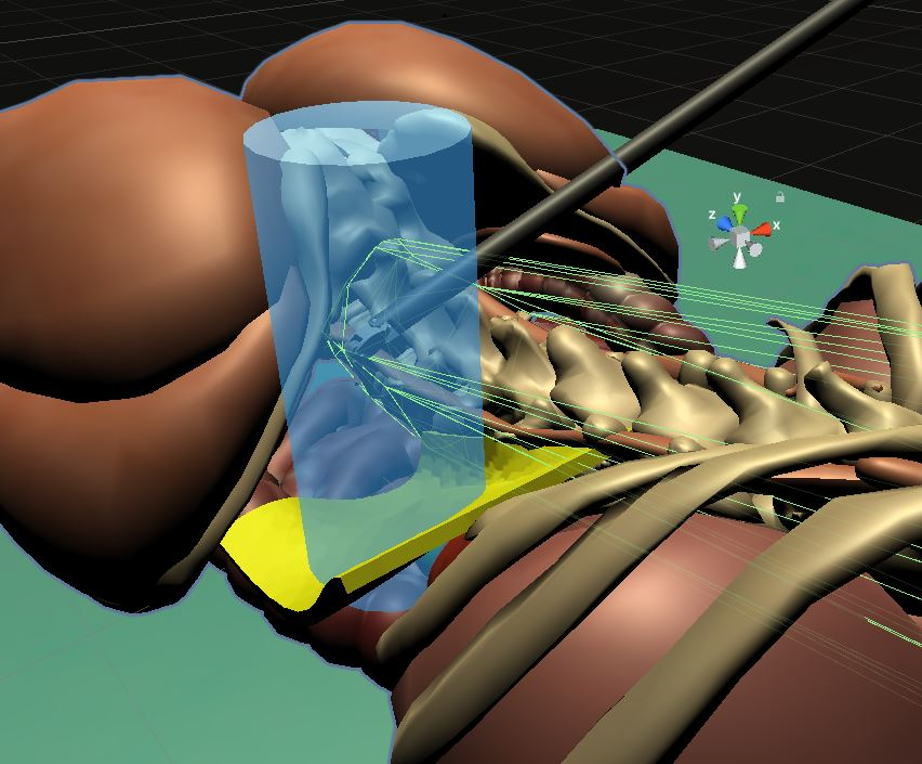}
	\caption{Explanatory overview of the safe end-effector workspace (light blue cylinder) and the mesh colliders (green lines) for the spinal column.}
	\label{fig:collision_scene}
\end{figure}

As depicted in Fig.~\ref{fig:collision_scene}, \textit{UnityFlexML} allows adding a mesh collider to our 3D organ models to automatically detect collisions between the PSM and anatomical organs, enabling reward shaping using this information. In more detail, a collision is triggered as an atomic event when an intersection between the bounds of two or more meshes is detected. Hence, by shaping the colliders for the different components of our training scene, we can detect such undesired collisions.

\subsection{Observation and Action Space} \label{obs_space}
Our DRL agent is represented by the End-Effector (EE) of the PSM, which interacts with the human body environment. The anatomical information such as the organs and tumour location is assumed to be known from pre-operative data. The TR task consists of moving the PSM from an initial position $p_0$, to the desired position $p_{tumour}$ in the proximity of the tumour and lift the fat tissue to a target location $p_{target}$, exposing the tumour. Note that the initial position of the PSM $p_0$ is randomised at the start of each episode during training. The state and action spaces of our environment are:

\begin{equation}
    \begin{split}
        S_t &= [g_t, p_t, p_i, \Vert p_t - p_i \Vert] \\
        A_t &= [\Delta_{t, j}]
    \end{split}
\end{equation}

\noindent where $g_t$ $\in$ $\{0, 1\}$ is the gripper state (open or close), $p_t$ is the position of the EE, $p_i$ is either $p_{tumor}$ in the first part of the trajectory (i.e., $g_t$ $=$ $0$) or $p_{target}$ in the lifting part (i.e., $g_t$ $=$ $1$), and $\Vert . \Vert$ is the Euclidean distance between the EE current position and the current target. In the action space, $\Delta_{t, j}$ $=$ $0.5\alpha$ (with $\alpha$ $\in$ $\{0, -1, +1\}$ controls the EE to move backward or forward by $0.5mm$ in the $j_{th}$ spacial dimension, or remain still.

\subsection{Reward Shaping} \label{sec:reward_shaping}
The gripper state $g_t$ is responsible for selecting the goal in the observation space of our agent; hence we design a reward function based on $g_t$ and the distance from the goal. Furthermore, we exploit the mesh collision system of \textit{UnityFlexML} to add a penalty term $c$ to the reward when the EE exits from the designed workspace (depicted as a light blue cylinder in Fig. \ref{fig:collision_scene}), or the PSM arm touches one of the organs:

\begin{equation}
    r(s_t) =
    \begin{cases}
        -(\Vert p_t - p_{tumour} \Vert \cdot k - 0.5) - c, \text{ if } g_t = 0\\
        -\Vert p_t - p_{target} \Vert \cdot k - c, \text{ if } g_t = 1
    \end{cases}
\end{equation}

\noindent where $k$ is a normalisation factor, 
and c is a constant penalty set to $1$ in case of collisions. Note that the scalar quantity of -0.5 is added to restrict the reward in the range [-1.0, -0.5] before grasping and [-0.5,0] after grasping. The reward trivially encourages the PSM towards the tumour when the gripper is open; otherwise, it favours movements toward the target position. 

\subsection{Training Algorithm}
We interfaced our \textit{UnityFlexML} environment with an external DRL software module written in Python to evaluate the performance of different DRL algorithms and choose the best performing one for the formal verification process. 
Among Twin Delayed Deep Deterministic Policy Gradient (TD3) \cite{TD3}, Soft Actor-Critic (SAC) \cite{haarnoja2018soft}, and Proximal Policy Optimisation (PPO) \cite{PPO}, we chose the latter one as it provided the best overall returns in our initial evaluation in terms of hyperparameter tuning and wall-clock training time. It is out of the scope of this work to obtain the best performance, while the main goal is to introduce and demonstrate the impact of safety constraints in DRL training. In more detail, we use the $\epsilon$-clipped implementation with $\epsilon$ $=$ $0.2$ as suggested in \cite{PPO}, to which we refer the interested reader for further details. 



\begin{figure}[t]
    \centering
    \vspace{.5em}
    \includegraphics[width=1\linewidth]{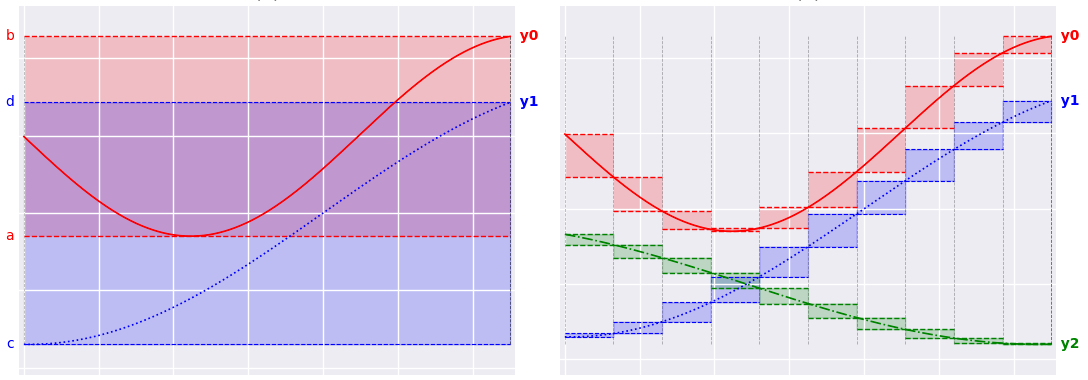}
    \caption{Explanatory output analysis of (left) decision-making problem with two outputs and one subdivision, and (right) output analysis with three outputs and multiple subdivisions}
    \vspace{-2mm}
    \label{fig:output_analysis}
\end{figure}

\subsection{Formal Analysis} \label{formal_analysis}

We define the formal verification framework such that given a set of properties, the framework should return whether the property is satisfied or provide counterexamples. We use the formalisation of Liu et al. \cite{verification_survey} for the safety properties that encodes an input-output relationship as:

\begin{equation} \label{eq:standard_proeprty}
\Theta:  x_0\in[a_0, b_0] \land ... \land x_n\in[a_n, b_n] \Rightarrow y_j\in[c, d]
\end{equation}
\noindent where $x_k \in X$ (i.e., input space), with $k \in [0, n]$, where n denotes the size of input states (i.e. dimension of $X$) and $y_j$ is a generic output of the network. Here, $ a_k, b_k, c, d \in \mathbb{R}$ represents the input and output bounds, respectively.

Such a property formulation is designed to verify if the network's output lies in a specific interval. 
However, in a DRL scenario, the network encodes a decision-making problem where each output node represents the value or the probability of a particular action. The agent selects the action with the highest probability or value with some stochasticity.
Therefore, we reformulate Proposition.~\ref{eq:standard_proeprty} to verify if one of the output values is lower than the others as follows:

\begin{equation} \label{eq:decision_proeprty}
\Theta: x_0\in[a_0, b_0] \land ... \land x_n\in[a_n, b_n] \Rightarrow y_j > y_i
\end{equation}

To verify the property, we rely on the Moore's comparison rules for intervals \cite{interval_algebra, moore1963interval}. In particular, assuming $y_i=[a, b]$ ($a, b \in \{a_k\},\{b_k\}$) and $y_j=[c, d]$, we obtain the proposition:
\begin{equation} \label{eq:moore_property} 
b < c  \Rightarrow y_i < y_j
\end{equation}

We exploit a layer-by-layer propagation to formally obtain an estimation of the output given an input interval \footnote{Project implementation: https://github.com/Ameyapores/SafeRLSurgery}. However, even in an ideal scenario where the estimated bounds perfectly match the real maximum and minimum values that the output nodes could assume (as in Fig.~\ref{fig:output_analysis} on the left), we can not formally guarantee if the property is respected. In the example, $y_1$ is lower than $y_0$ in the entire input domain (x-axis), but considering the estimated bound limits $y_0=[a, b]$ and $y_1=[c, d]$, we can not formally conclude if the decision-making property is proved or denied using Proposition.~\ref{eq:moore_property}, because $d \nless a$. To summarising, formal verification based on Proposition.~\ref{eq:moore_property}, only considers the estimated bound limits to verify a property; hence these approaches typically fail at directly verifying properties on large input domains.
To solve this problem, we propose to subdivide the input domain of the property in a set of sub-intervals (\textit{subarea}) and analyse them independently. Fig.~\ref{fig:output_analysis}(right) shows this process, where the sub-intervals allows to obtain a better estimation of the output function's shapes (and bounds). Hence it allows a straightforward application of the Moore rules for the interval comparison. Notice that a situation such as Fig.~\ref{fig:output_analysis}(left) could still happen in a certain \textit{subarea}. We can also address this by recursively iterating our process until $d < a$ (property verified) or $c > b$ (property violated) on that particular \textit{subarea}. In the right Fig.~\ref{fig:output_analysis}, the property $y_1 < y_0$ is proved in the whole domain, while the property $y_1 < y_2$ is clearly violated in the second half of the input domain.
This formulation is one of the first efforts in adapting formal analysis techniques to a reinforcement learning problem.

\subsection{Violation Rate}
In this section, we introduce a novel metrics based on our formal verification approach to evaluate the safety of a trained model with respect to a set of safety properties. One of the limitations of the classical verification algorithms is that they only provide yes (if the property is respected in the input domain) or no (if the property is violated in one point). In contrast, we propose to compute the percentage of the domain violations of the desired properties to evaluate the model safety. Intuitively, we keep track of the subarea size at each iteration of our method that violates the properties.
At the end of the process, we get a \textit{violation rate} which is the subarea size normalised over the initial size of the domain.
The violation rate represents an upper bound of the probability to violate the safety properties.

\section{EXPERIMENTS} \label{experiments}
The task of tissue retraction is divided into two phases: approaching the tumour and retracting the fat tissue once it is grasped. Our objective is to demonstrate that by using safety criteria, i.e. the collision penalty, the overall safety of the surgical procedure increases. For this, we define a safe workspace for both the subtasks such that there is no collision of the PSM with the surrounding tissue inside the workspace. Fig.~\ref{fig:collision_scene} shows the safe workspace for the approaching phase. This replicates the surgical scenario where we avoid collisions with hard anatomical parts such as the ribs and spinal column that can cause severe consequences, whereas we ignore the collision with soft tissues present near the region of interest. For each workspace, we define safety properties, i.e. an upper bound and a lower bound, such that the configuration of the PSM satisfying the property is considered safe. In the approach phase, we define properties for each direction (in the Cartesian space), i.e. $\Theta_{1R}, \Theta_{1L}$ represent the left and right constraints in the x-direction. Similarly, $\Theta_{2R}$, and $\Theta_{3R},  \Theta_{3L}$ represents the constraints in the y and z directions, respectively. Please note that, for the approaching phase, we do not consider any upper limit on the y-axis $\Theta_{2L}$ since there are no obstacles in that direction. We define similar properties, $\Theta_{4R}- \Theta_{6L}$ for the retract phase. 
We report a detailed description of all the proposed properties in Table.~\ref{tab:violation}.
We train PPO considering these safety properties (Safe-PPO) by penalising the agent when it violates these properties, i.e. it goes out of the safe workspace as described in Sec.~\ref{sec:reward_shaping}. 

In our experiments, firstly, we compare the performance of Safe-PPO in achieving high rewards with PPO that does not consider safety constraints (Unsafe-PPO). We show the violation rate of all properties using formal analysis for both the Safe-PPO and Unsafe-PPO. 
Moreover, to identify the contribution of the considered properties towards the overall behaviour, we provide an ablation study in which we train PPO that considers a subset of properties and compute the violation rate. Table.~\ref{tab1:brain_definition} illustrates the selected policies that are trained considering various properties. 

\begin{table}[h!]
\centering
\caption{Considered policies used in the ablation study.}
\label{tab1:brain_definition}
\begin{tabular}{p{0.27\linewidth} p{0.63\linewidth}}
\hline\noalign{\smallskip}
Brains    &   Properties used for training \\ 
\noalign{\smallskip}\hline\noalign{\smallskip}
Safe-PPO &  All properties ($\Theta_{1R}-\Theta_{6L}$)\\
Unsafe-PPO & No properties\\
Primitive Safe-PPO & Safe-PPO in early stages (after 400 epochs) of the training ($\Theta_{1R}-\Theta_{6L})$)\\
Policy4 & First set of properties ($\Theta_{1R},\Theta_{1L},\Theta_{2R}$)\\
Policy5 & Second block of properties ($\Theta_{3R},\Theta_{3L},\Theta_{4R}$)\\
Policy6 & Last set of properties ($\Theta_{4L},\Theta_{5R},\Theta_{5L}, \Theta_{6R}, \Theta_{6L}$)\\
\noalign{\smallskip}\hline
\end{tabular}
\end{table}

Next, using the trained model for Safe-PPO, we investigate the possibility of knowing a priori the input states that can lead to unsafe configurations. The formal verification tool segments the input domain into sub-intervals and recursively iterates the splitting with different heuristics until it can prove (or disprove) the violation criteria for each sub-interval (see Sec.~\ref{formal_analysis}). This enables us to identify all the state values of the considered inputs that can cause violation for the DRL policy. To understand whether a standard execution using Safe-PPO encounters states that cause violation, we analyse the inputs for 1000 episodes and visualise the state distribution.  

Additionally, we estimate the ability of the learnt behaviour in exposing the tumour using a tumour exposure metrics (TE). TE computes the normalised percentage of the tumour surface that can be seen from a camera positioned in front of the region of interest. Following \cite{tagliabue2020soft}, the safe workspace is divided into a grid of 5x5 aligned with the x-z plane (see Fig.~\ref{fig:collision_scene})
The EE position is initiated at each point in the grid and the number of pixels of the tumour is computed through the camera. 
This evaluation allows us to evaluate the impact of the safety constraints in the tumour exposure while varying the initial position of EE.



\section{RESULTS AND DISCUSSION} \label{resutls}
\begin{figure}[t]
    \vspace{.5em}
    \centering
    \includegraphics[width=.9\linewidth]{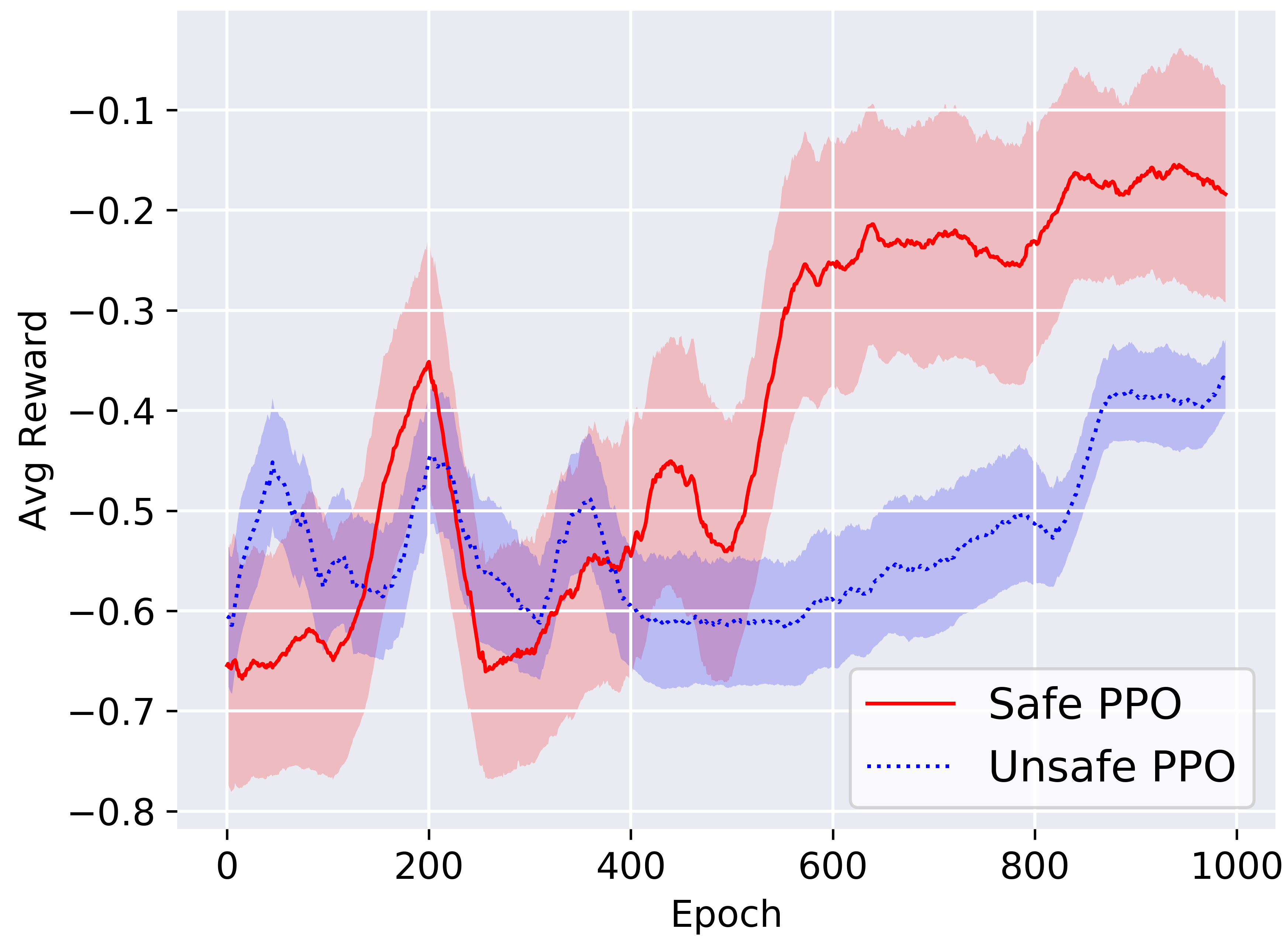}
    \caption{The obtained learning curves for Safe-PPO and Unsafe PPO. The curves are averaged over four different seeds and smoothed over 25 epochs.}
    \label{fig:reward_plot}
\end{figure}
The two policies, i.e. Safe-PPO and Unsafe-PPO, learn the complete task in approx 800 epochs, where each epoch is 2000 time steps. Fig.~\ref{fig:reward_plot} depicts the average reward achieved as a function of training steps. Both policies learn the first phase of approaching the lesion fast; however, Safe-PPO incurs collision penalty at the start, which unsafe does not. Hence, Safe-PPO stays lower in reward compared to Unsafe-PPO until 400 epochs. After 400 epochs, Safe-PPO correctly learns the trajectories to approach the lesion avoiding unsafe configurations. Hence, it reaches a higher reward after 400 epochs. The rewards further show a drastic increase at 800 epochs. This rise can be associated with learning safe trajectories for the retract phase.

\begin{figure}[t]
    \centering
    \includegraphics[width=1\linewidth]{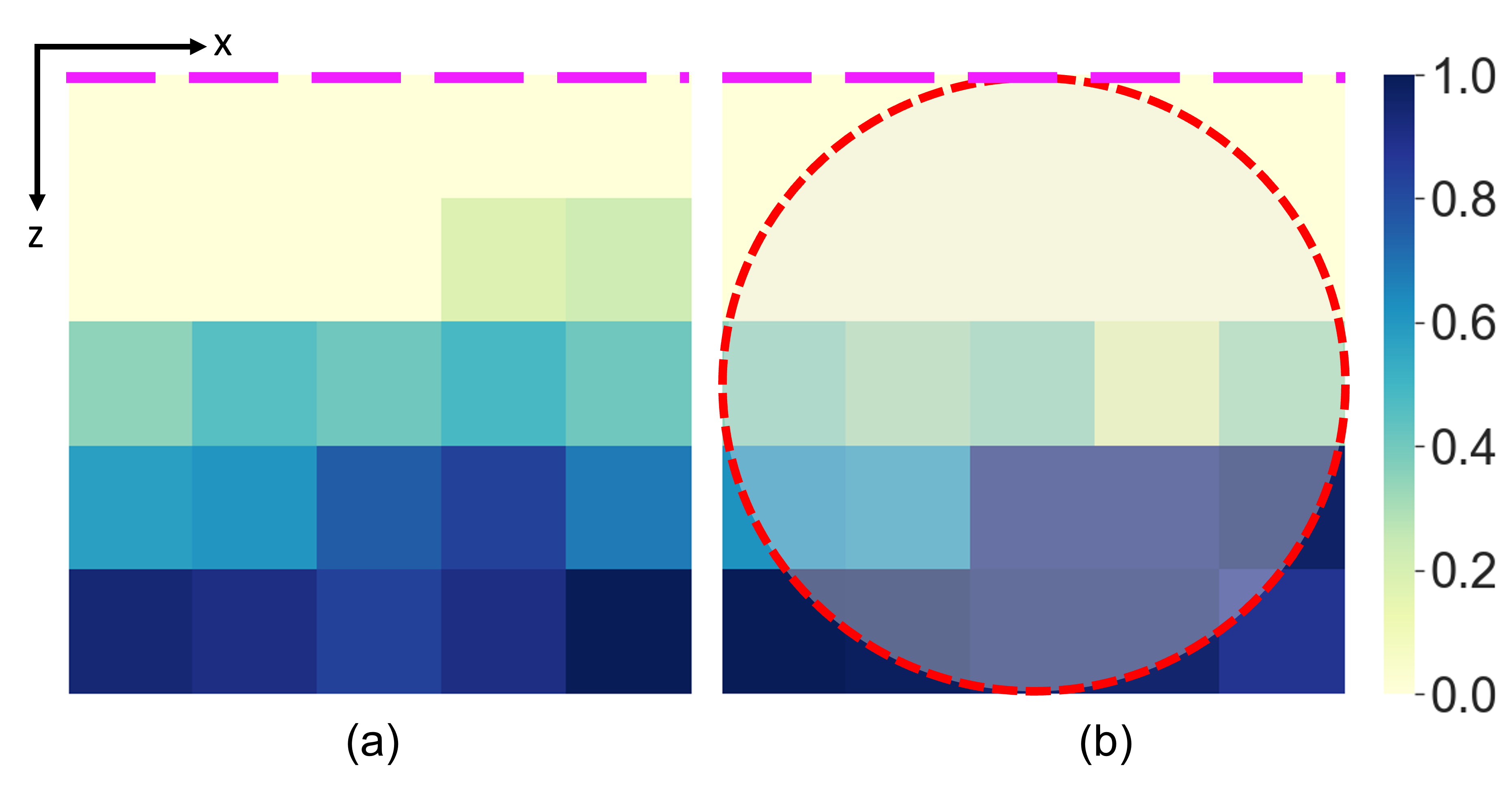}
    \caption{
    Top-view of safe end-effector workspace showing the Tumour Exposure (TE) from different starting points (a) Unsafe-PPO (b) Safe-PPO. The marked circle shows the safe workspace projection, while the dashed pink line represents the attachment region. See text for more details.}
    \label{fig:heatplot}
\end{figure}
\begin{table*}
\centering
\caption{SUMMARY OF VIOLATION RATES FOR EACH PROPERTY} \label{tab:violation}
\begin{tabular}{ l l l l p{0.10\linewidth} l l l } \toprule
& & \multicolumn{6}{c}{Violation rate (\%)} \\
\noalign{\smallskip }\cline{3-8} \noalign{\smallskip} 
\textbf{Properties}    & \textbf{Property description} & \textbf{Safe-PPO} & \textbf{Unsafe-PPO} & \textbf{Primitive Safe-PPO} & \textbf{Policy4} & \textbf{Policy5} & \textbf{Policy6}\\ \toprule
$\Theta_{1L}$ & Lower limit on x-direction (approach) & 24.4 & 91.4 & 32.0 & 0.0 & 14.0 & 86.7\\
$\Theta_{1R}$ & Upper limit on x-direction (approach) & 0.0 & 7.6 & 57.6 & 0.0 & 51.2 & 7.0\\
$\Theta_{2L}$ & Lower limit on y-direction (approach) & 0.0 & 0.0 & 0.0 & 0.0 & 0.0 & 0.0 \\
$\Theta_{3L}$ & Lower limit on z-direction (approach) & 9.4 & 61.5 & 0.0 & 0.0 & 14.3 & 40.0\\
$\Theta_{3R}$ & Upper limit on z-direction (approach) & 0.0 & 29.8 & 100.0 & 100.0 & 0.0 & 32.3\\
$\Theta_{4L}$ & Lower limit on x-direction (retract) & 0.0 & 11.3 & 14.1 & 10.5 & 100.0 & 1.6\\
$\Theta_{4R}$ & Upper limit on x-direction (retract) & 0.0 & 20.7 & 100.0 & 0.0 & 0.0 & 0.0\\
$\Theta_{5L}$ & Lower limit on y-direction (retract) & 0.0 & 75.0 & 81.2 & 0.0 & 0.0 & 0.0\\
$\Theta_{5R}$ & Upper limit on y-direction (retract) & 0.0 & 0.0 & 43.7 & 49.2 & 0.0 & 0.0\\
$\Theta_{6L}$ & Lower limit on z-direction (retract) & 0.0 & 0.0 & 3.7 & 0.0 & 36.7 & 0.0\\
$\Theta_{6R}$ & Upper limit on z-direction (retract) & 0.0 & 0.0 & 0.0 & 0.0 & 0.0 & 0.0 \\ \midrule
Average ($\Theta_{1L}$-$\Theta_{2L}$) & & 8.14 & 33.01 & 29.88 & 0.0 & 21.74 & 31.25\\
Average ($\Theta_{3L}$-$\Theta_{4R}$) & & 4.69 & 45.61 & 50.00 & 50.00 & 7.13 & 36.15\\ 
Average ($\Theta_{5L}$-$\Theta_{6R}$) & & 0.00 & 17.84 & 40.46 & 9.96 & 22.79 & 0.26\\ \midrule
Overall Average & & \textbf{3.07} & 27.02 & 39.31 & 14.52 & 19.65 & 15.24\\
\bottomrule
\end{tabular}
\end{table*}

Table.~\ref{tab:violation} shows the results of the violation rate for the ablation studies using formal verification. Several policies are trained as listed in Table.~\ref{tab1:brain_definition}. 
Safe-PPO has a mean global violation rate of 3.07\%, whereas Unsafe-PPO shows a mean violation rate of 27\%. This indicates that incorporating safety criteria through a collision penalty can drastically increase the safety of the procedure. 
However, reporting the average violation rate does not provide an intuition of the distribution as some properties tend to be more important than others in terms of the damage that can be caused should they be violated. For example,
a large proportion of Safe-PPO violation corresponds to $\Theta_{1L}$. This suggests that $\Theta_{1L}$ is a difficult property to satisfy due to the presence of a complex obstacle and corresponds to the presence of the spinal column in that direction.
As previously mentioned, during early training, Primitive Safe-PPO incurs several collision penalties and hence initially hinder the overall safety of the trajectory (Refer to the third column in Table.~\ref{tab:violation}). 
Therefore, it shows a higher violation rate for many properties.
Furthermore, we can infer that all the policies stay within the safety limits for properties $\Theta_{2L}, \Theta_{6R}$, hence their violation rate remains 0\%. 
Policy4, Policy5 and Policy6 consider a subset of properties while training, and their average global violation rate remains between that of Safe-PPO and Unsafe PPO. These policies show 0\% violations on the properties that are considered for their training but a significantly higher violation rate on other properties. One reason for this could be compensatory behaviour in which optimising for one set of properties leads to unsafe configurations on some other properties; however, further investigation is required to explain high violation rates on certain properties.

\begin{figure}[t]
    \centering
    \includegraphics[width=1\linewidth]{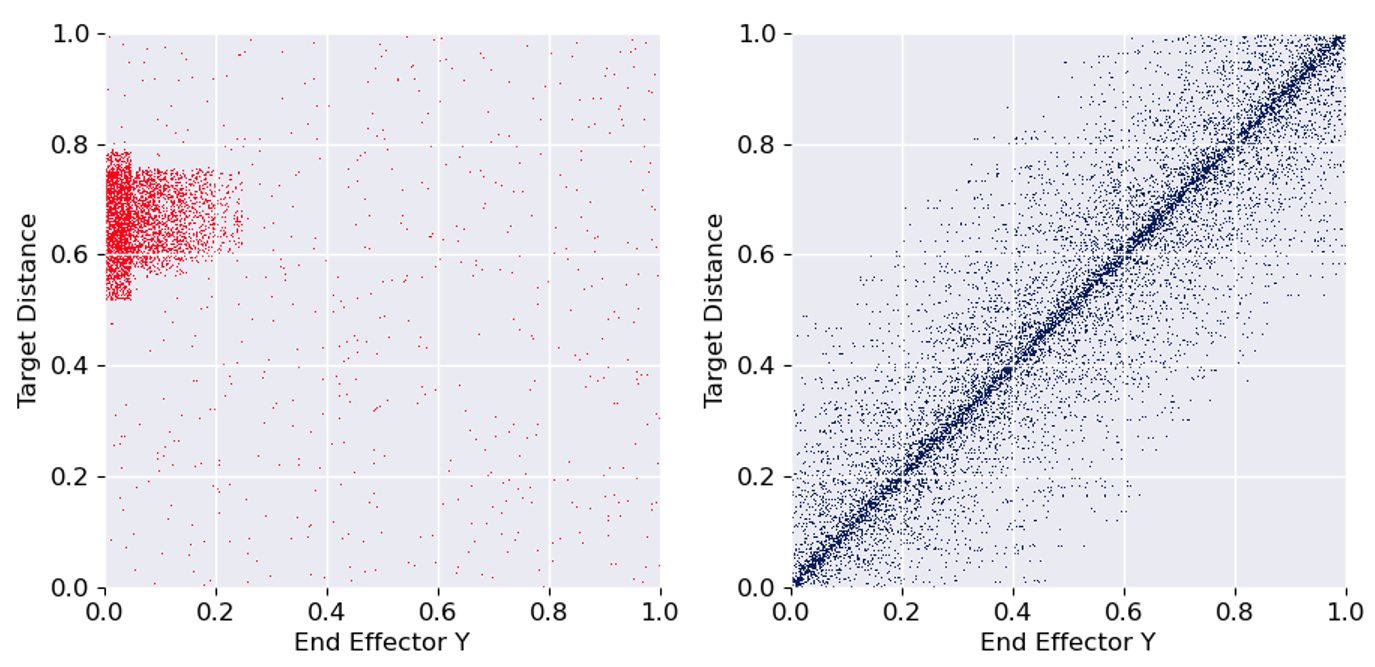}
    \caption{(left) State values that cause a violation for Safe-PPO derived using the formal verification tool and (right) State distribution in a standard execution of Safe-PPO (1000 episodes). We describe the relationship between two-state inputs, i.e. normalised EE movement in the y-direction and target distance, to simplify visualisation and use static values for other inputs.}
    \label{fig:violation_rate}
\end{figure}

For Safe-PPO, we represent the states that can cause violation using the formal verification tool in Fig.~\ref{fig:violation_rate}(left), whereas the states that are frequently encountered in standard execution in Fig.~\ref{fig:violation_rate}(right). 
Since the observation space consists of a continuous 7-dimensional input and a discrete input for grasping (see Sec.~\ref{obs_space}), to visualise the states in 2D, we fix the values for the x and z motion of the end-effector (by sampling from a normal distribution for each episode) and apply the formal verification tool for the entire input domain of the EE movement in the y-direction and the target distance respectively.
Fig.~\ref{fig:violation_rate}(right) indicates a linear relationship between EE movement in the y-direction and the target distance. 
Fig.~\ref{fig:violation_rate}(left) shows that most state violations occur for lower values of EE Y in the range [0.0,0.2] and higher values of target distance in the range [0.5,0.8].
These violations are non-intuitive because they can be caused due to violations in other state inputs since they are normally sampled. 
Fig.~\ref{fig:violation_rate} shows that Safe-PPO rarely encounter states that cause a violation.
Even if in a real-robotic system, such adversarial perturbations can occur rarely, they can cause fatal consequences. 
Thanks to the proposed formal verification tool, we can identify these rare, occurring hazardous states in advance by the policy.
Note that incorporating safety constraints inside the training loop does not add to any additional costs in terms of computational time. The formal analysis is an offline process performed after the training and does not influence the learned behaviour.
We show the obtained TE matrix in Fig.~\ref{fig:heatplot} for Unsafe-PPO and Safe-PPO. The two methods show similar TE in all considered grid locations. The average TE for Unsafe-PPO and Safe-PPO is almost identical, 0.42 and 0.41, respectively. This indicates that adding safety conditions does not affect the overall task performance. Hence, our method (Safe-PPO) provides a safety guarantee with optimal performance with respect to TE. 
Note that the TE is low in the proximal region of the fat tissue attachment corresponding to the upper area of the plots in Fig.~\ref{fig:heatplot}.
In this region, the EE grasps the fat tissue near the attachment and reaches the target position without any TE. 
One explanation for this is because the reward function changes drastically upon grasping the fat and does not penalise if the grasping point is far from the tumour. 
In the regions distal from the attachment, the grasping point comes closer to the tumour, consequently exposing the tumour. We plan to introduce the TE factor in the reward function for future research activities to improve this behaviour.

\section{CONCLUSIONS} \label{conclusion}
In this work, we address the risks associated with the actions during DRL training, especially for safety-critical scenarios such as surgical robotics. We propose a Safe-DRL framework in which safety constraints could be added through reward shaping. Further, we formulate a formal verification tool to evaluate the violations caused by a DRL policy. This tool allows us to identify the states that can cause safety violations a priori to model execution.
In this study, we automate the task of TR that is commonly performed in multiple phases of MIS in a virtual environment. The risks associated with TR consists of surrounding tissue damage if the robotic end-effector violates the workspace limits. Therefore, we design a safe workspace and add safety criteria for violating the workspace. Our results indicate an increased safety and more reliable trajectories performed using the safety protocol than DRL methods without safety. 

Our future work will include implementing the formal verification controller online during the execution to prevent undesirable actions. 
Subsequently, we plan to migrate the simulation experiments to synthetic phantom studies on a real robotic system using the pipeline established earlier \cite{tagliabue2020soft}. Moreover, further investigation is required to transfer the learnt policies to different scenes with similar surgical gestures.
A critical derivative that emerges from this study is that some properties are more important than others. Giving weights for different properties using prior knowledge of the surgical scenario can potentially lead to safer behaviour.




\bibliographystyle{IEEEtran}
\bibliography{root}

\end{document}